\documentclass[10pt,twocolumn,letterpaper]{article}

\usepackage{wacv}
\usepackage{times}
\usepackage{epsfig}
\usepackage{graphicx}
\usepackage{amsmath}
\usepackage{amssymb}

\usepackage{color,soul}  
\definecolor{ao(english)}{rgb}{0.0, 0.5, 0.0}

\def\etal{\emph{et al.\ }}

\wacvfinalcopy 

\ifwacvfinal
\def\assignedStartPage{1} 
\fi

\ifwacvfinal
\usepackage[breaklinks=true,bookmarks=false]{hyperref}
\else
\usepackage[pagebackref=true,breaklinks=true,colorlinks,bookmarks=false]{hyperref}
\fi

\ifwacvfinal
\else
\fi

\begin{document}

\title{VideoPose: Estimating $6$D object pose from videos}

\author{Apoorva Beedu$ ^*$\\
Georgia Institute of Technology\\
{\tt\small abeedu3@gatech.edu}
\and
Zhile Ren\\
Georgia Institute of Technology\\
{\tt\small  jrenzhile@gmail.com}

\and
Varun Agrawal\\
Georgia Institute of Technology\\
{\tt\small  varunagrawal@gatech.edu}

\and
Irfan Essa\\
Georgia Institute of Technology\\
{\tt\small  irfan@gatech.edu}
}
\maketitle
\begin{abstract}
We introduce a simple yet effective algorithm that uses convolutional neural networks to directly estimate object poses from videos. 
Our approach leverages the temporal information from a video sequence, and is computationally efficient and robust to support robotic and AR domains. Our proposed network takes a pre-trained 2D object detector as input, and aggregates visual features through a recurrent neural network to make predictions at each frame.
Experimental evaluation on the YCB-Video dataset show that our approach is on par with the state-of-the-art algorithms. Further, with a speed of 30 fps, it is also more efficient than the state-of-the-art, and therefore applicable to a variety of applications that require real-time object pose estimation. 


\end{abstract}

\section{Introduction}
\label{sec:intro}
Estimating the $3$D translation and $3$D rotation for every object in an image is a core building block for many applications in robotics~\cite{saxena2008robotic,Tremblay2018DeepOP,eppner2019billion} and augmented reality~\cite{marchand2015pose}.
The classical solution for such $6$-DOF pose estimation problems utilises a feature point matching mechanism, followed by Perspective-n-Point (PnP) to correct the estimated pose~\cite{Rad2017BB8AS,tekin2018real,peng2019pvnet,hu2019segmentation}.
However, such approaches fail when objects are texture-less or heavily occluded. Typical ways of refining the $6$DOF estimation involves using additional depth data~\cite{wang2019densefusion,Brachmann2014Learning,hinterstoisser2012model,konishi2018real} or post-processing methods like ICP or other deep learning based methods ~\cite{xiang2018posecnn,Kehl2017SSD6DMR,li2018deepim,song2020hybridpose}, which increase computational costs. Other approaches treat it as a classification problem~\cite{tulsiani2015viewpoints,Kehl2017SSD6DMR}, resulting in reduced performance as the output space is not continuous. 


In robotics, augmented reality, and mobile applications, the input signals are usually videos rather than a single image. Li~\etal\cite{li2018unified} utilize multiple frames from different viewing angles to estimate single object poses, which does not work robustly in complex scenes. Wen~\etal\cite{wen2020se} and Deng~\etal\cite{ deng2019poserbpf} use tracking methods to estimate the poses, however these methods do not explicitly exploit the temporal information in the videos. The idea of using more than one frame to estimate object poses has seen limited exploration. As the object poses in a video sequence are implicitly related to camera transformations and do not change abruptly between frames, and as different viewpoints of the objects aid in the pose estimation ~\cite{labbe2020cosypose, chen2017multi}, we believe that modelling a temporal relationship can only help the task. 

Motivated by this, in our proposed approach, we utilize a simple CNN-based architecture to extract useful features, and subsequently aggregate the information across consecutive frames using a recurrent neural network (RNN). 
The training is performed on the YCB-Video dataset~\cite{xiang2018posecnn} and the approach achieves comparable performance to state-of-the-art approaches, while requiring lower computational costs.~We also conduct extensive ablation studies and demonstrate the effectiveness of our network design. 

The primary contributions of our paper are:

\begin{itemize}
    \item We introduce a simple yet effective neural network architecture for estimating $6$-DOF object poses from videos. Our system first extracts image features and estimates depth and labels, then use a temporal module to aggregate information across frames and estimate $6$-DOF pose of every object in the current frame. 
    \item We perform extensive ablation studies on different design choices of the system, and show that using videos, as opposed to using single images, can improve the predictions significantly at an improved computational speed of 30 fps.
    \end{itemize}

\begin{figure*}[]
    \centering
    \includegraphics[width=0.8\textwidth]{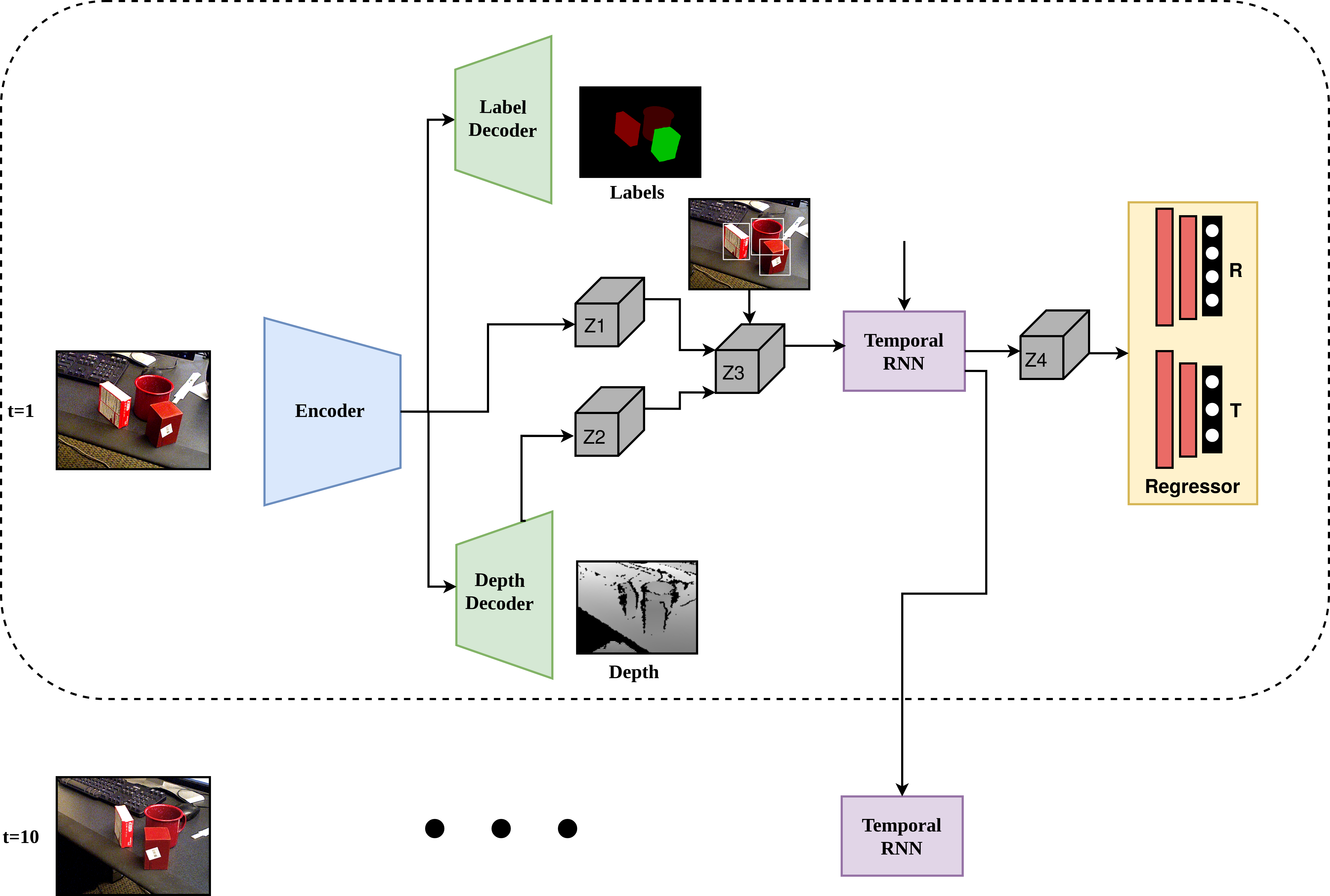} \vspace{.1in} 
    \caption{Overview of our VideoPose framework for 6D object pose estimation. We use the same encoder as in ~\cite{xiang2018posecnn}. $Z4$ is the fused features from Fig ~\ref{fig:simpleRNN} }
    \label{fig:overview}
\end{figure*}

\section{Related Work}
\label{sec:related_work}
Estimating the $6$-DOF pose of objects in the scene is a widely studied task. The classical methods either use template-based or feature-based approaches. In template-based methods, a template is matched to different locations in the image, and a similarity score is computed~\cite{hinterstoisser2012model,hinterstoisser2011gradient}. However, these template matching methods could fail to make predictions for textureless objects and cluttered environments. In feature based methods, local features are extracted, and correspondence between known 3D objects and local 2D features is established using PnP to recover 6D poses. However, these methods also require sufficient textures on the object to compute local features and face difficulty in generalising well to new environments as they are often trained on small datasets. 


Convolutional Neural Networks (CNNs) have proven to be an effective tool in many computer vision tasks. However, they rely heavily on the availability of large-scale annotated datasets. Thus, the YCB-Video dataset~\cite{xiang2018posecnn}, T-LESS~\cite{Hodan2017TLESSAR}, and OccludedLINEMOD dataset~\cite{krull2015learning,peng2019pvnet} were introduced. These datasets have enabled the emergence of novel network designs such as PoseCNN, DPOD~\cite{zakharov2019dpod} and PVNet~\cite{xiang2018posecnn,peng2019pvnet}. To further increase the amount of accurate annotated data, Trembly~\etal~\cite{tremblay2018falling} introduced synthetically generated photo-realistic data, which when trained on, gave improved performances on the estimation task~\cite{Tremblay2018DeepOP}. In this paper, we use the challenging YCB-Video dataset, as it is a popular dataset that serves as a testbed for many recent algorithms. 


Building on those datasets, various CNN architectures have been introduced to learn effective representations of objects, and thus estimate accurate 6D poses. Kehl~\etal~\cite{Kehl2017SSD6DMR} extends SSD~\cite{liu2016ssd} by adding an additional viewpoint classification branch to the network. Rad~\etal~\cite{Rad2017BB8AS} and Telkin~\etal~\cite{tekin2018real} predict 2D projections from 3D bounding box estimations. However, these methods fail to deal with pose ambiguities and objects under heavy occlusion. 
Most notably, PoseCNN~\cite{xiang2018posecnn} uses a Hough voting scheme to vote for the center of the object and then use the bounding boxes to estimate the $3$D rotation.
To address the problems of heavy occlusions and ambiguities,\cite{peng2019pvnet,hu2019segmentation,park2019pix2pose,oberweger2018making} learn to detect keypoints and then perform PnP.  However, these methods also encounter similar problems of pose ambiguities for symmetric and partially occluded objects.

Other methods involve a hybrid approach where the model learns to perform multiple tasks. Song~\etal~\cite{song2020hybridpose} enforce consistencies among keypoints, edges, and object symmetries. Billings~\etal~\cite{billings2019silhonet} predict silhouettes of objects along with object poses. There is also a growing trend of designing model agnostic features~\cite{wang2019normalized} that can handle novel objects. These directions are beyond the scope of our paper, as our goal is to find the best practice for pose estimation in videos when the objects of interest are given.


\begin{figure*}[ht!]
    \centering
    \includegraphics[width=0.8\textwidth]{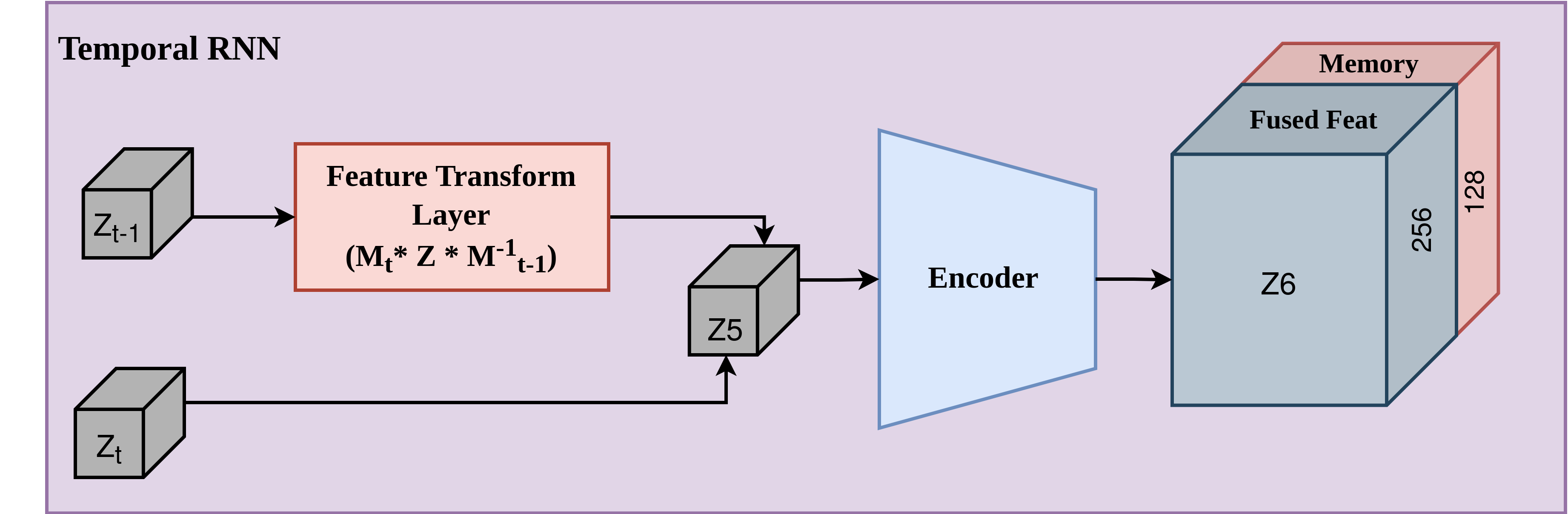}
    \caption{Overview of a simple temporal RNN network: $Z_{t-1}$ is the feature from the previous time-step and $Z_t$ is the $Z3$ from Fig~\ref{fig:overview}.}
    \label{fig:simpleRNN}
\end{figure*}

To refine the predicted poses, several works use additional depth information and perform a standard ICP algorithm~\cite{xiang2018posecnn,Kehl2017SSD6DMR}, or directly learn from RGB-D inputs~\cite{wang2019densefusion,li2018deepim,zakharov2019dpod}. 
We argue that since the input signals to robots and/or mobile devices are typically video sequences, instead of heavily relying on additional depth information, estimating poses in videos by exploiting the temporal data could already refine the single pose estimations. 
A notable work from Deng~\etal~\cite{deng2019poserbpf} introduces the PoseRBPF algorithm that uses particle filters to track objects in video sequences. 
This state-of-the-art algorithm provides accurate estimations at a high computational cost. 
Wen~\etal~\cite{wen2020se} also perform tracking, but use synthetic rendering of the object at the previous time-step. 
In contrast to the above papers, we show that a simple temporal module that aggregates information across different frames at a high computational speed performs comparable or better than using single frames. 


\section{Approach}
\label{sec:methodology}
Given an RGB video stream, our goal is to estimate the $3$D rotation and $3$D translation of all the objects in every frame of the video. 
We assume the system has access to the $3$D model of the object. In the following sections, $\boldsymbol{R}$ denotes the rotation matrix with respect to the annotated canonical object pose, and $\boldsymbol{T}$ is the translation from the object to the camera.

\subsection{Overview of the network}
\label{sec: overview of the network}
Our pipeline, shown in Figure \ref{fig:overview} consists of two stages. The first stage corresponds to the feature extractor, depth estimator and semantic label predictor. The second has a temporal RNN and a Regressor.
For extracting image features, we use a simple VGG-$16$ model similar to ~\cite{xiang2018posecnn}, and fine-tune the last $2$ layers to encode features from the depth and semantic prediction tasks. 

Pose Estimation relies on accurate object detection. The object detection module is responsible for giving the class-id and Region-Of-Interest (ROI). During training, we use the ground truth bounding box and during testing, we use the predictions and bounding box from the PoseCNN model. This can ideally be replaced with any lightweight feature extraction model such as  MobileNet~\cite{howard2017mobilenets} to make the inference faster, but we choose bounding boxes from PoseCNN for fair comparison to prior works. 
We learn the depth and semantic segmentation using a Decoder consisting of 3 CNN layers. The final layer from the feature extractor and the penultimate layer of the depth estimator are concatenated and pooled together by using ROI Align \cite{he2017mask} which is passed through the temporal layer. We believe that adding depth features can aid in the estimation, when depth information is not available.


\subsection{Temporal block}
To use the features from the previous step, we project the image features from the previous step to current step using the camera transformation matrix $M$ as $M_t * \text{feat} * M^{-1}_{t-1}$, which is subsequently concatenated with the features from the current time-step.
This is passed to an Encoder network and its output is divided into two parts -- the first 128 layers representing the memory, and the remaining 256 layers representing the fused features. 

The fused features are further passed through a regressor module to estimate the poses, and the memory features are used for the next time-step.  We perform the transformation based on the ground truth camera transformations. 
This is illustrated in Figure \ref{fig:simpleRNN}.
\label{sec: overview of simpleRNN}


\begin{table*}[ht]
\centering
\resizebox{0.8\textwidth}{!}{%
\begin{tabular}{|c|c|c|c|c|c|c|c|c|c|c|}
\hline
{ } &  \multicolumn{2}{l|}{PoseCNN} &
  \multicolumn{2}{l|}{\begin{tabular}[c]{@{}l@{}}PoseRBPF \\ (50 particles)\end{tabular}} &
  \multicolumn{2}{l|}{\begin{tabular}[c]{@{}l@{}}VideoPose \\ (GT BBox)\end{tabular}} &
  \multicolumn{2}{l|}{\begin{tabular}[c]{@{}l@{}}VideoPose \\ (PoseCNN BBox)\end{tabular}} \\ \hline
{ } & ADD & ADD-S & ADD & ADD-S & { ADD} & { ADD-S} & { ADD} & { ADD-S} \\ \hline
002\_master\_chef\_can & 50.9 & 84 & \textbf{56.1} & 75.6 & 55.5 & 85.0 &
  \underline{52.1} &
  \textcolor{ao(english)}{\underline{\textbf{84.3}}} \\ \hline
003\_cracker\_box &
  51.7 & 76.9 & \textbf{73.4} & \textbf{85.2} & 10.9 & 63.3 & 6.9 &
  58.6 \\ \hline
004\_sugar\_box &  68.6 & 84.3 & \textbf{73.9} &  \textbf{86.5} &
  47.1 &  71.9 &  41.2 &  68.6 \\ \hline
005\_tomato\_soup\_can & 66 & 80.9 & \textbf{71.1} &
  82.0 &   62.6 &   83.3 &   61.0 &
  \textcolor{ao(english)}{\underline{\textbf{83.2}}} \\ \hline
006\_mustard\_bottle &
  79.9 &   \textbf{90.2} &   \textbf{80.0} &
  90.0 &   67.9 &   85.9 &   73.7 & 88.7 \\ \hline
007\_tuna\_fish\_can &
  \textbf{70.4} & \textbf{87.9} & 56.1 & 73.8 & 56.1 & 83.3 & 53.1 &
  \textcolor{ao(english)}{82.10} \\ \hline
008\_pudding\_box &
  \textbf{62.9} & \textbf{79} & 54.8 & 69.2 &  56.7 &  76.6 & 48.9 &
  \textcolor{ao(english)}{71.32} \\ \hline
009\_gelatin\_box &
  75.2 &  87.1 & \textbf{83.1} &  \textbf{89.7} & 76 & 87.2 & 70.8 & 84.6 \\ \hline
010\_potted\_meat\_can &
  \textbf{59.6} &
  \textbf{78.5} &   47.0 &   61.3 &   45.9 & 77.7 &  41.4 & \textcolor{ao(english)}{75.6} \\ \hline
011\_banana &
  \textbf{72.3} &   \textbf{85.9} & 22.8 &  64.2 &  40.6 &  70.7 &  \textcolor{ao(english)}{43.4} &
  \textcolor{ao(english)}{72.1} \\ \hline
019\_pitcher\_base &
  {52.5} &  {76.8} &
  {\textbf{74.0}} &  {\textbf{87.5}} &   60 &   82.5 &  48.8 &   \underline{77.30} \\ \hline
021\_bleach\_cleanser &
  {50.5} &
  {\textbf{71.9}} &  {\textbf{51.6}} & {66.7} &  41 & 59.9 &  29.5 &  50.4 \\ \hline
024\_bowl &
  {6.5} &
  {69.7} &
  {\textbf{26.4}} &
  {\textbf{88.2}} &
  1.5 &
  73.2 &
  1.6 &
  67.1 \\ \hline
025\_mug &
  {57.7} &
  {78} &
  {\textbf{67.3}} &
  {\textbf{83.7}} &
  56.3 &
  85.4 &
  43.2 &
  77.9 \\ \hline
035\_power\_drill &
  {55.1} &
  {75.8} &
  {\textbf{64.4}} &
  {\textbf{80.6}} &
  26.4 &
  63.9 &
  16.5 &
  61.9 \\ \hline
036\_wood\_block &
  {\textbf{31.8}} &
  {\textbf{65.8}} &
  {0.0} &
  {0.0} &
  0.00 &
  16.30 &
  0.00 &
  \textcolor{ao(english)}{13.6} \\ \hline
037\_scissors &
  {\textbf{35.8}} &
  {56.2} &
  {20.6} &
  {30.9} &
  29.5 &
  62.5 &
  \textcolor{ao(english)}{27.9} &
  \textcolor{ao(english)}{\underline{\textbf{72.1}}} \\ \hline
040\_large\_marker &
  {\textbf{58}} &
  {\textbf{71.4}} &
  {45.7} &
  {54.1} &
  21.6 &
  55.6 &
  20.5 &
  \textcolor{ao(english)}{54.2} \\ \hline
051\_large\_clamp &
  {25} &
  {49.9} &
  {\textbf{27.0}} &
  {\textbf{73.3}} &
  14 &
  61.5 &
  16.5 &
  \underline{55.3} \\ \hline
052\_extra\_large\_clamp &
  {15.8} &
  {47} &
  {\textbf{50.4}} &
  {\textbf{68.7}} &
  55.3 &
  49.9 &
  4.1 &
  46.0 \\ \hline
061\_foam\_brick &
  {40.4} &
  {87.8} &
  {\textbf{75.8}} &
  {\textbf{88.4}} &
  43.1 &
  80.7 &
  \underline{40.9} &
  77.5 \\ \hline \hline
ALL &
  {53.7} &
  {\textbf{75.9}} &
  {\textbf{57.1}} &
  {74.8} &
  39.7 &
  71.2 &
  35.4 &
  68.3 \\ \hline
\end{tabular}
}
\hspace{2em}
\vspace{.1in}
\caption{Quantitative Evaluation of 6D poses (ADD and ADD-S) on the YCB-Video Dataset. \textbf{Bold} values compare between PoseCNN, PoseRBPF and VideoPose using the PoseCNN bbox. Values in \textcolor{ao(english)}{green} compares the columns with PoseRBPF, underlined values compare with PoseCNN} 
\label{table:results_on_ycb}
\end{table*}
\subsection{6D Pose Regresssion}
The translation vector $\boldsymbol{T}$ is the object location in the camera coordinate system. A naive way of estimating $\boldsymbol{T}$ is to directly regress to it. However, doing so cannot handle multiple object instances or generalise well to new objects. To tackle this problem, Xiang~\etal~\cite{xiang2018posecnn} estimate $\boldsymbol{T}$ by localising the $2$D object center in the image and estimating object distance from the camera. Suppose \begin{math}\mathbf{c} = (c_x, c_y)^T \end{math}  are the centers of the object in the frame and $T_z$ is either learnt or estimated from the depth image, then \textit{$T_x$} and \textit{$T_y$} can be estimated as:
\begin{gather}
 \begin{bmatrix}c_x \\ c_y \end{bmatrix}
 =
 \begin{bmatrix}
  f_x\displaystyle\frac{T_x}{T_z}+ p_x \\
  f_y\displaystyle\frac{T_y}{T_z} + p_y
  \end{bmatrix}, \label{eq:1}
\end{gather} where \textit{$f_x$} and \textit{$f_y$} are focal lengths and $(\textit{$p_x$}, \textit{$p_y$})^\textit{T}$ are principal points. Since we have rough estimates of object locations from the noisy object detection inputs, we train our model to estimate \begin{math}\Delta c_x, \Delta c_y, \text{and} \ T_z \end{math}. We then estimate \textit{$T_x$} and \textit{$T_y$} using the following equation:
\begin{gather}
 \begin{bmatrix}c_x + \Delta c_x \\ c_y + \Delta c_y\end{bmatrix}
 =
 \begin{bmatrix}
  f_x\displaystyle\frac{T_x}{T_z}+ p_x \\
  f_y\displaystyle\frac{T_y}{T_z} + p_y
  \end{bmatrix}. \label{eq:2}
\end{gather}

The fused features from the temporal module are then fed to 2 disconnected regressor blocks - a $2$ layer FCN Regressor module, with 512 dimensions and $3 \times n$ where $n$ is the number of objects for the translation and a $2$ layer FCN regressor  the temporal features are fed to a regressor with 512 dimensions and $4 \times$. We disconnect $T_x , T_y$ and $T_z$ by training two separate linear layers to account for the different dimensions learnt.
Similar to \cite{xiang2018posecnn}, we represent the rotation $\boldsymbol{R}$ using quaternions.

\subsection{Training Strategy}
We use the $L_1$ loss to learn depth ($L_\text{depth}$), and cross entropy loss for semantic segmentation ($L_\text{label}$). 
The pose estimation loss is obtained by projecting the $3$D points using the estimated and ground truth pose, and then computing their distance:
\begin{equation}
L_\text{pose}(\mathbf{\widetilde{q}}, \mathbf{q}) = \frac{1}{m} \sum\limits_{x \in M} || (R(\mathbf{\widetilde{q}})x + \mathbf{\widetilde{t}}) - (R(\textbf{q})x + \mathbf{t})|| ^2,
\end{equation} where $M$ denotes the set of 3D points, $m$ is total number of points. $R(\mathbf{\widetilde{q}})$ and $R(\mathbf{q})$ indicate the rotation matrix computed from the quaternion representation as in \cite{xiang2018posecnn}.
In addition, we also add a cosine loss on the quaternions, and  regularisation loss to force the norm of the quaternion to be 1. Quaternions that represent rotations are unit norm, and forcing the norm to be bounded by 1 helps in the learning process by reducing the scope. 
\begin{equation}\label{eq:lreg}
    L_\text{reg} = ||1 - \text{norm}(\mathbf{\widetilde{q}}) ||,~~~       
    L_\text{inner\_prod} = 1 - \langle\mathbf{\widetilde{q}},\textbf{q}\rangle.
\end{equation}
The total loss can be defined as 
\begin{equation} \label{eq:loss2}
    L(\mathbf{\widetilde{q}}, \mathbf{q}, \mathbf{\widetilde{t}}, \mathbf{t}) = L_\text{depth} + L_\text{label} + L_\text{pose} + L_\text{reg} + L_\text{inner\_prod}.
\end{equation}

\section{Experiments} 
\label{sec:experiments}
Now we compare our method with PoseCNN\cite{xiang2018posecnn} and PoseRBPF\cite{deng2019poserbpf}. We also conduct ablation studies on the choice of model architecture, and the number of video frames the model requires to perform well. 
\subsection{Dataset}
We evaluate the proposed method on the YCB-Video dataset \cite{xiang2018posecnn}. Details of which are explained in sec \ref{sec:implementation}. \textbf{The YCB-Video Dataset} contains 92 RGB-D video sequences of 21 objects. The dataset contains textured and textureless objects of varying shape, and different levels of occlusion where about $15\%$ of objects are heavily occluded. Objects are annotated with 6D poses, segmentation masks and depth images. For our purposes, We create smaller video sequences of 10 RGB images by taking every alternate frame from the video. \\

\subsection{Metrics}
We use two metrics to report on the YCB-Dataset. ADD is the average distance between the corresponding points on the 3D object at the ground truth and predicted poses. Given the estimated $[\mathbf{\widetilde{R}}|\mathbf{\widetilde{t}}]$ and the ground truth poses $[\mathbf{R}|\mathbf{t}]$, ADD-S, designed for symmetric objects, calculates the mean distance from each $3$D point to a closest point on the target model. 

\subsection{Implementation}
\label{sec:implementation}
VideoPose is implemented using the PyTorch \cite{paszke2017automatic} framework. We use a learning rate of $5e^{-4}$ and the Adam optimiser~\cite{kingma2014adam} with a weight decay of $1e^{-5}$.
Learning rate is multiplied by $0.8$ after every $5$ epochs, until it hits a lower bound of $1e^{-6}$. 
For the feature encoder, we freeze the VGG$16$ weights from PoseCNN, and train rest of the network from scratch. 
We create video samples of 10 frames and train our model for ~100 epochs with the learning schedule described above.

During training, we augment the input images with colour-jitter and noise, and for the bounding box, we augment it by extending the height and width randomly between $0$ and $10\%$ of the height and width of the object. 
While training the temporal block, we create videos with random time jumps in between. For instance, given a large video sequence, we create video samples $1:n:10*n$, where n is a random number between $1$ and $10$, thus forcing the model to account for small and large jumps between consecutive frames.

\subsection{Evaluation}
\label{sec:result}
We compare our results with PoseCNN \cite{xiang2018posecnn} for single frame prediction and PoseRBPF \cite{deng2019poserbpf} for videos in Table \ref{table:results_on_ycb}. We compare two different approaches for computing the ROI: $1)$ Using the ground truth; $2)$ Using the ROI predicted by PoseCNN. 
In order to maintain comparable FPS, the PoseRBPF is computed using 50 particles.

\begin{figure*}[]
\centering
\begin{minipage}[c]{0.5\textwidth}
\centering
    \includegraphics[width=1\textwidth]{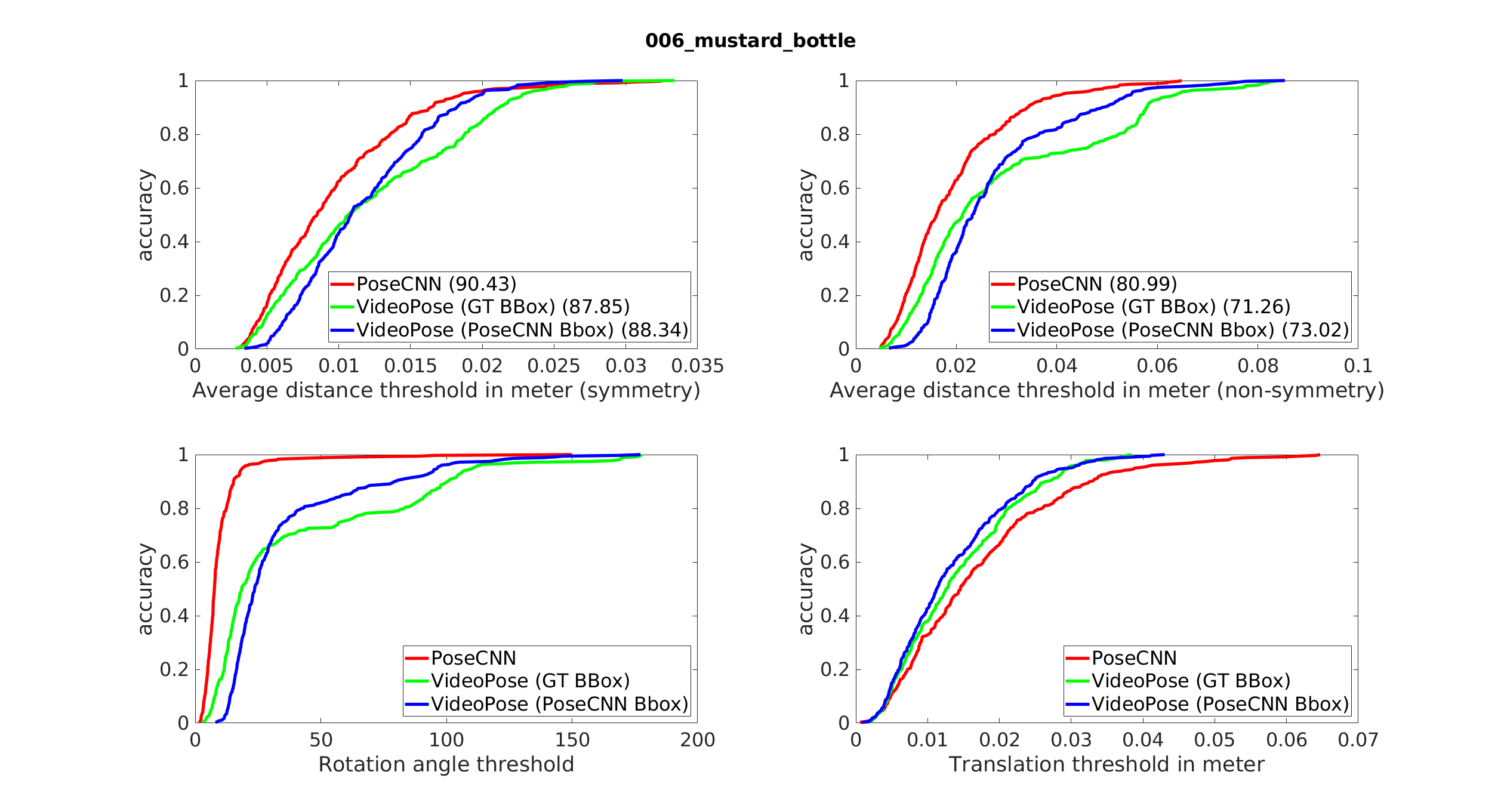}
\end{minipage}
\begin{minipage}[c]{0.495\textwidth}
\centering
    \includegraphics[width=1\textwidth]{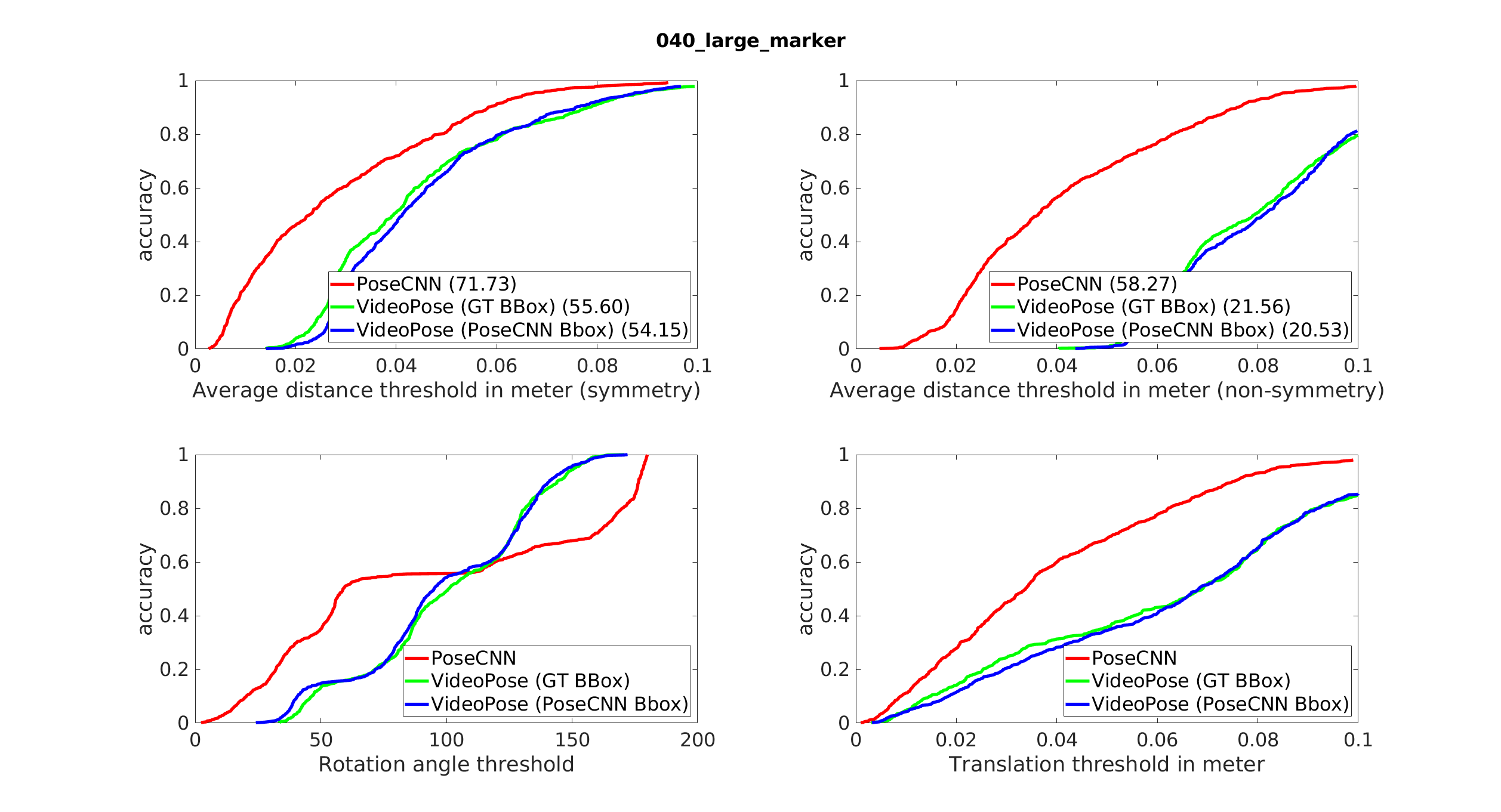}
\end{minipage}

\centering
\begin{minipage}[c]{0.5\textwidth}
\centering
    \includegraphics[width=1\textwidth]{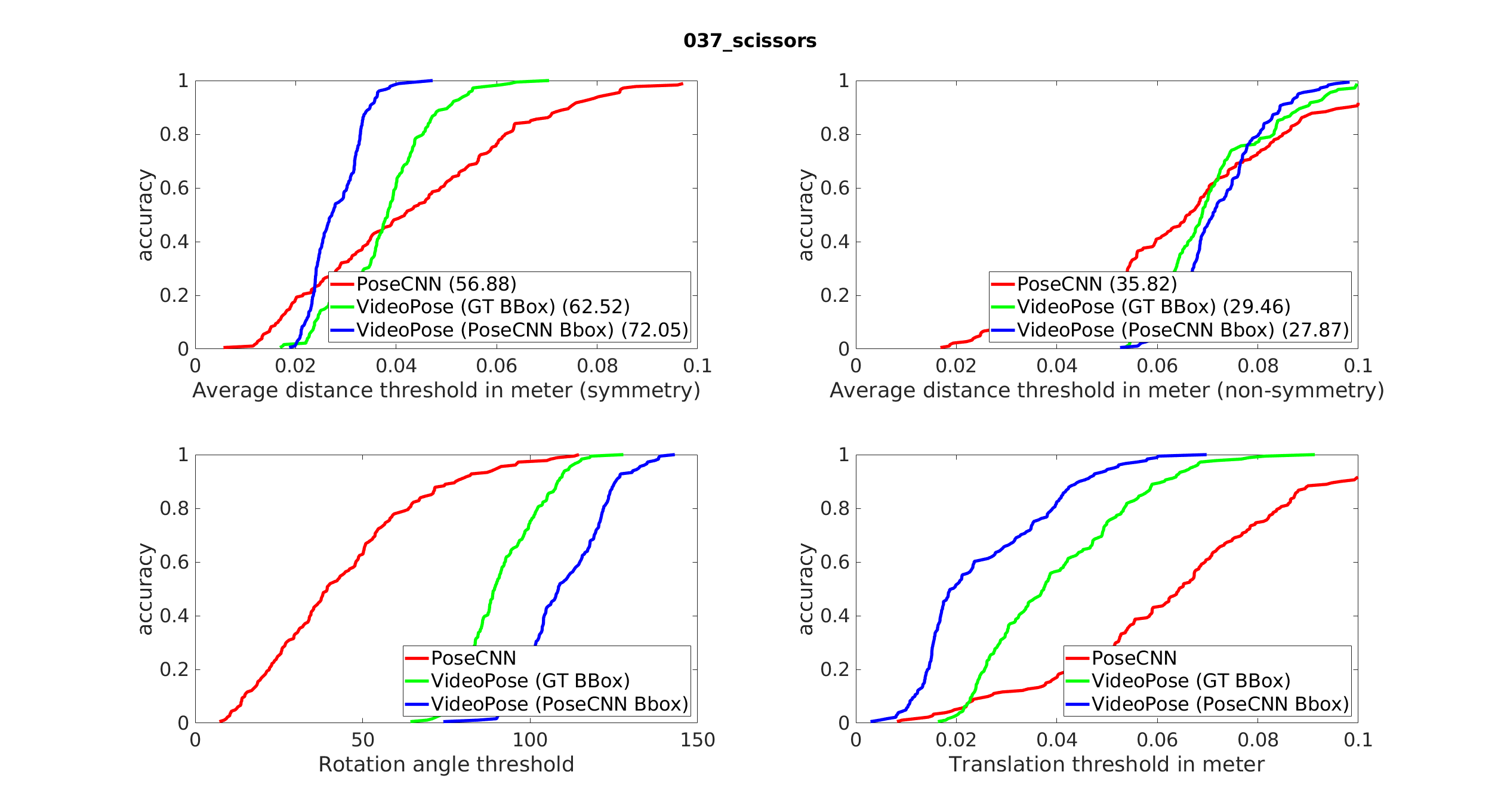}
\end{minipage}
\begin{minipage}[c]{0.495\textwidth}
\centering
    \includegraphics[width=1\textwidth]{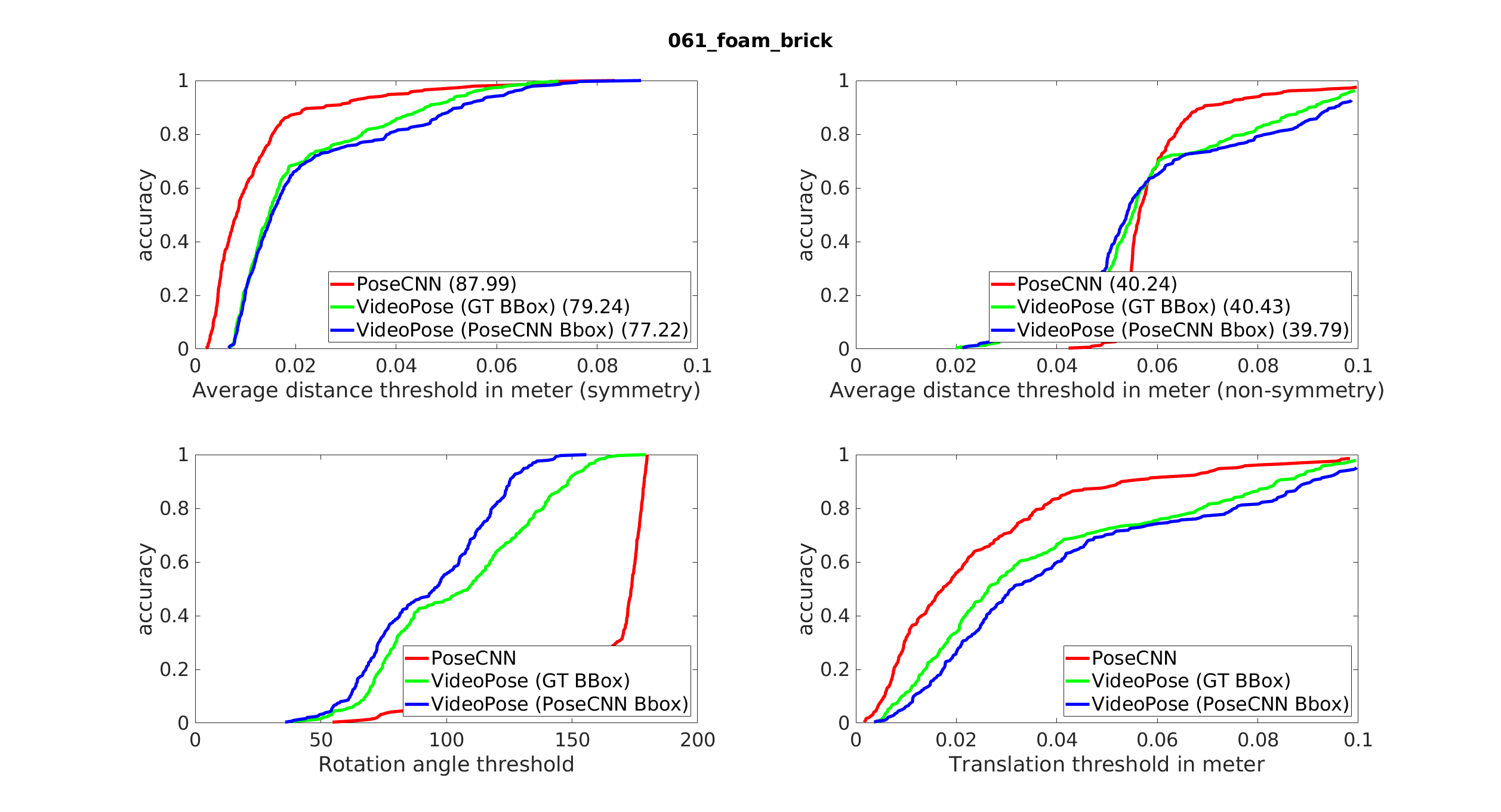}
\end{minipage}
\caption{AUC with rotation and translation curves for few objects}
\label{fig:auc}
\end{figure*}

We observe that, VideoPose, when using the bounding boxes from PoseCNN, has a small drop in the accuracy, showing that our method is robust to noises in the ROIs. For objects where our method is not comparable to the sota, we further look into the AUC curves in Fig. ~\ref{fig:auc} and note that our method outperforms PoseCNN in rotation for symmetric objects like foam\_brick and large\_marker, and in translation for scissors and mustard\_bottle which are non-symmetric. 

\begin{table}[ht]
\centering
\resizebox{1\columnwidth}{!}{%
\begin{tabular}{|c|c|c|c|c|c|c|c|c|}
\hline
{ }
{ } &  \multicolumn{2}{l|}{PoseCNN} &
  \multicolumn{2}{l|}{\begin{tabular}[c]{@{}l@{}}VideoPose\\ (ConvGRU)\end{tabular}} &
  \multicolumn{2}{l|}{\begin{tabular}[c]{@{}l@{}}VideoPose \\ (baseline)\end{tabular}} \\ \hline
 { } & ADD & ADD-S & { ADD} & { ADD-S} & { ADD} & { ADD-S} \\ \hline
002\_master\_chef\_can &
  50.9 &
  84 &
  36.34 &
  80.86 &
  \textbf{55.5} &
  \textbf{85.0}
  \\ \hline
003\_cracker\_box &
  \textbf{51.7} &
  \textbf{76.9} &
  22.51 &
  62.14 &
  10.9 &
  63.3
  \\ \hline
004\_sugar\_box &
  \textbf{68.6} &
  \textbf{84.3} &
  40.73 &
  68.76 &
  47.1 &
  71.9
  \\ \hline
005\_tomato\_soup\_can &
  66 &
  80.9 &
  \textbf{66.99} &
  \textbf{83.49} &
  62.6 &
  83.3
  \\ \hline
006\_mustard\_bottle &
  \textbf{79.9} &
  \textbf{90.2} &
  75.33 &
  88.96 &
  67.9 &
  85.9
  \\ \hline
007\_tuna\_fish\_can &
  \textbf{70.4} &
  \textbf{87.9} &
  60.58 &
  86.35 &
  56.1 &
  83.3
  \\ \hline
008\_pudding\_box &
  \textbf{62.9} &
  \textbf{79} &
  49.56 &
  75.77 &
  56.7 &
  76.6
  \\ \hline
009\_gelatin\_box &
  75.2 &
  87.1 &
  \textbf{81.06} &
  \textbf{89.32} &
  76 &
  87.2
  \\ \hline
010\_potted\_meat\_can &
  59.6 &
  78.5 &
  \textbf{61.54} &
  \textbf{83.64} &
  45.9 &
  77.7
  \\ \hline
011\_banana &
  \textbf{72.3} &
  \textbf{85.9} &
  22.31 &
  69.59 &
  40.6 &
  70.7
  \\ \hline
019\_pitcher\_base &
  {52.5} &
  {76.8} &
  {\textbf{70.54}} &
  {\textbf{85.92}} &
  60 &
  82.5
  \\ \hline
021\_bleach\_cleanser &
  {\textbf{50.5}} &
  {\textbf{71.9}} &
  {46.93} &
  {62.11} &
  41 &
  59.9
  \\ \hline
024\_bowl &
  {6.5} &
  {69.7} &
  {\textbf{12.78}} &
  {\textbf{80.08}} &
  1.5 &
  73.2
  \\ \hline
025\_mug &
  {57.7} &
  {78} &
  {\textbf{67.62}} &
  {\textbf{88.79}} &
  56.3 &
  85.4
  \\ \hline
035\_power\_drill &
  {\textbf{55.1}} &
  {\textbf{75.8}} &
  {35.99} &
  {71.21} &
  26.4 &
  63.9
  \\ \hline
036\_wood\_block &
  {\textbf{31.8}} &
  {\textbf{65.8}} &
  {0.00} &
  {28.72} &
  0.00 &
  16.30
  \\ \hline
037\_scissors &
  {35.8} &
  {56.2} &
  {\textbf{50.08}} &
  {\textbf{73.39}} &
  29.5 &
  62.5
  \\ \hline
040\_large\_marker &
  {\textbf{58}} &
  {\textbf{71.4}} &
  {36.81} &
  {53.89} &
  21.6 &
  55.6
  \\ \hline
051\_large\_clamp &
  {\textbf{25}} &
  {49.9} &
  {20.72} &
  {\textbf{69.33}} &
  14 &
  61.5
  \\ \hline
052\_extra\_large\_clamp &
  {\textbf{15.8}} &
  {47} &
  {5.93} &
  {\textbf{55.39}} &
  5.3 &
  49.9
  \\ \hline
061\_foam\_brick &
  {40.4} &
  {\textbf{87.8}} &
  {\textbf{44.78}} &
  {86.14} &
  43.1 &
  80.7
  \\ \hline
ALL &
  {\textbf{53.7}} &
  {\textbf{75.9}} &
  {44.09} &
  {73.90} &
  39.7 &
  71.2
  \\ \hline
\end{tabular}
}
\hspace{2em}
\vspace{.1in}
\caption{Comparison of performance between different architectures. \textbf{Bold} values represent the best method for a given object.}
\label{table:diff_arch}
\vspace{.1in} 
\end{table}

\begin{table}[b]
\centering
\scalebox{0.775}{%
\begin{tabular}{|c|c|c|c|c|c|}
\hline
Methods & ~\cite{xiang2018posecnn} & ~\cite{deng2019poserbpf} (50) & ~\cite{deng2019poserbpf} (200) & Ours(RNN) & Ours(ConvGRU)\\ \hline
Time (fps) & 5.88 & 20 & 5 & 30 & 25\\ \hline
\end{tabular}
}
\vspace{.05in} 
\caption{Comparison of frame rates for different methods: PoseCNN~\cite{xiang2018posecnn}, PoseRBPF (50 particles)~\cite{deng2019poserbpf}, PoseRBPF (200 particles)~\cite{deng2019poserbpf}, VideoPose (baseline) and VideoPose(ConvGRU)}
\label{table:time_efficiency}
\vspace{.1in} 

\centering
\scalebox{0.75}{%
\begin{tabular}{|c|c|c|c|c|c|}
\hline
Methods  & Position=2 & Position=5 & Position=10 & Position=15 & Position=19\\ \hline
ADD & 41.6 & 41.61 & 44.08 & 41.3457 & 40.79\\ \hline
ADD-S & 71.77 & 71.83 & 73.9 & 71.71 & 71.18 \\ \hline
\end{tabular}
}
\vspace{.1in} 
\caption{Studying the number of previous frames required for a good estimate. Position refers to the location of the keyframe in a video sequence of 20 frames.} 
\label{table:comparison_for_different_keyframe_position}
\end{table}
\noindent\textbf{Impact of different temporal blocks} We also perform ablation studies on different architectures used to capture the temporality, as shown in Table \ref{table:diff_arch}. 
Instead of the baseline temporal RNN in fig. ~\ref{fig:simpleRNN}, we use a ConvGRU ~\cite{ballas2015delving} as the temporal module and observe that it handles the temporal information more effectively. The performance is comparable to or better than PoseCNN and PoseRBPF for more than 50\% of the objects. 
We treat this ablation study as a proof that using previous estimates can aid in the pose estimation, regardless of the temporal module used.

\noindent\textbf{Time efficiency} We compare the time efficiency of our model with the baseline models. 
The run time for PoseCNN is taken from Wang \etal~\cite{wang2019densefusion}. From Table \ref{table:time_efficiency}, we see that VideoPose with ConvGRU, is slightly faster than PoseRBPF while providing an increase in the performance. As the baseline temporal RNN is more lightweight than ConvGRUs, we get about 50\% increase in the speed compared to PoseRBPF, while maintaining the accuracy. \\

\noindent \textbf{Qualitative Analysis of the $6$D predictions} We show three examples of the predictions by VideoPose, PoseCNN, and ground truth poses in Table \ref{fig:qualitative1} \ref{fig:qualitative2}. The columns represent the $2$D projections of predictions using VideoPose, PoseCNN, and the ground truth poses, and rows specify the time-steps. More results are shown in the supplementary. We notice that the poses between frames are more consistent when estimated through videos as opposed to single images. We also observe that the initial frame estimation is as critical for our approach, as it is for other refinement methods.\\

\noindent\textbf{Effect of number of previous frames used} Table \ref{table:comparison_for_different_keyframe_position} shows the effect of number of previous frames used. We see that we get the best performance at position 10, and it reduces a little for later positions. It is worth noting that the model was trained for a video sequence of length 10. So the little performance drop for $15th$ and $19th$ position shows that our model, even when trained for a video sequence of 10, can effectively model longer sequences. This is due to the ablation of the video samples during training as discussed in ~\ref{sec:implementation}.

\begin{table*}[!ht]
\centering
\resizebox{1\textwidth}{!}{%
\scalebox{1}{%
  \begin{tabular}{|c|c|}
    \hline
     & VideoPose \hspace{2in} PoseCNN \hspace{2in}  GT \\ \hline
    t = 1 & \includegraphics[width=1.1\textwidth]{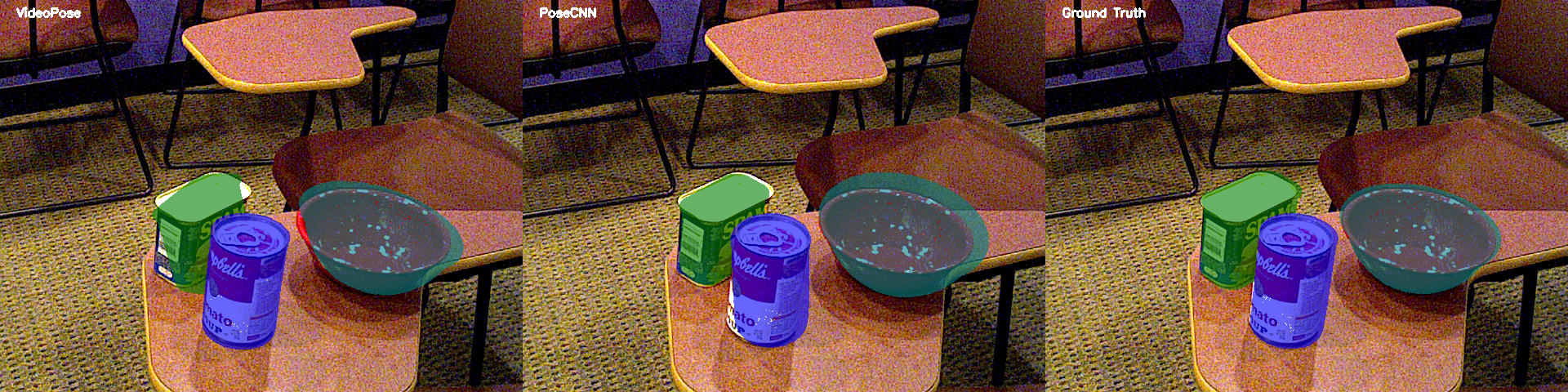} \\ \hline
    t = 2 & \includegraphics[width=1.1\textwidth]{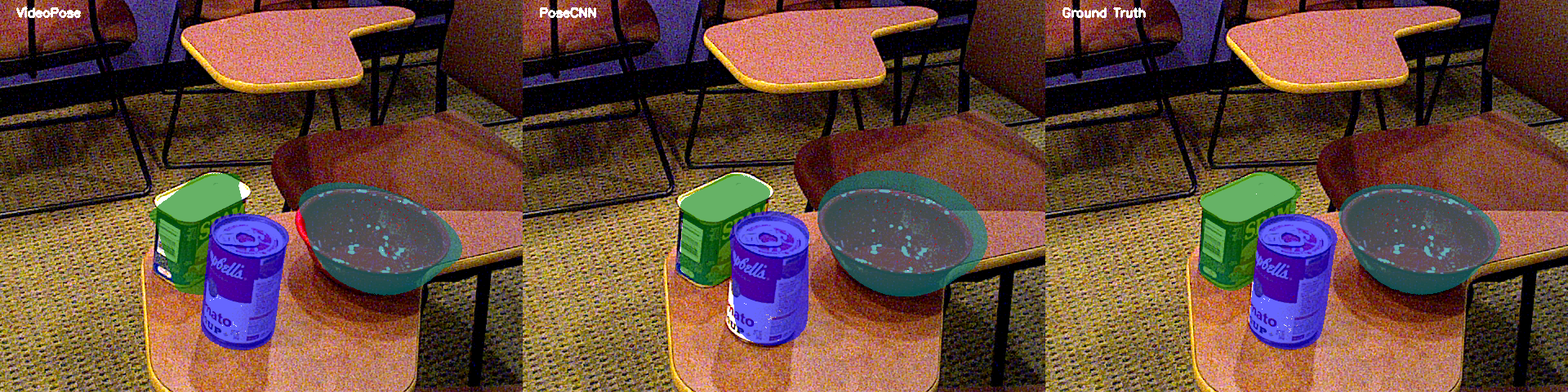} \\  \hline
    t = 3 & \includegraphics[width=1.1\textwidth]{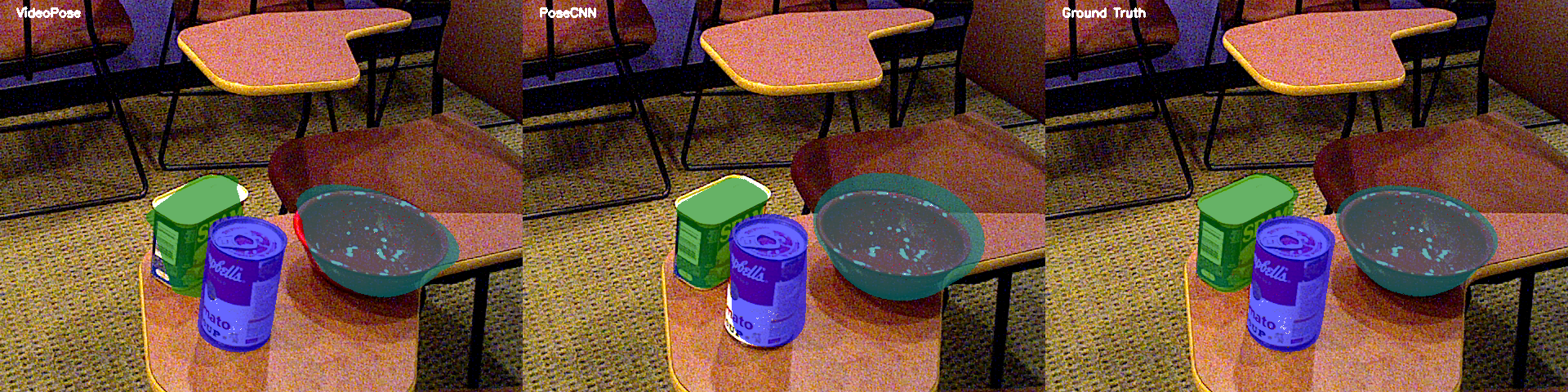} \\ \hline
    t = 4 & \includegraphics[width=1.1\textwidth]{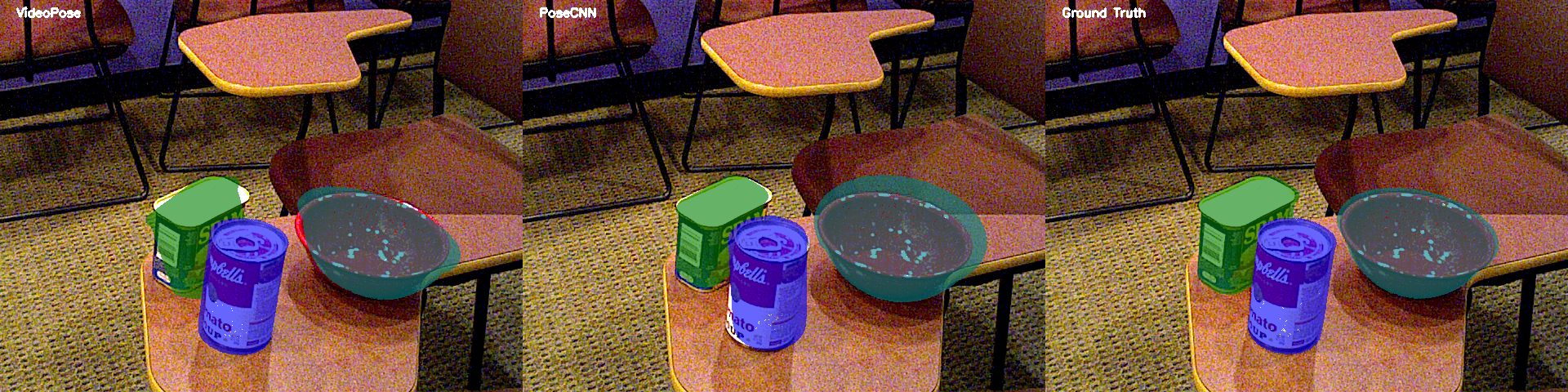} \\ \hline
    t = 5 & \includegraphics[width=1.1\textwidth]{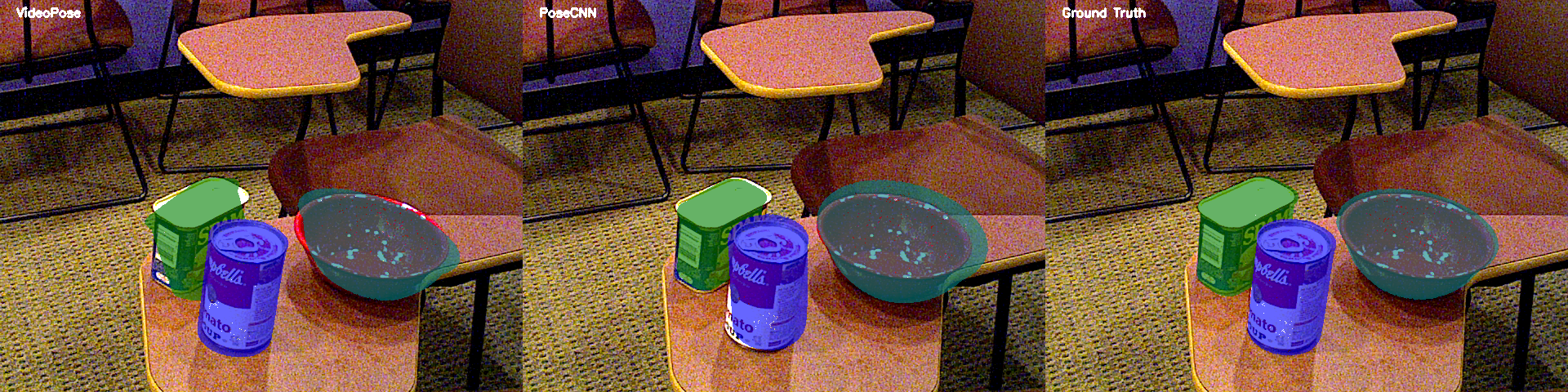}\\ \hline
  \end{tabular}
  }
  }
  \vspace{.1in}
  \caption{Visualisations of the estimated poses on YCB-Dataset for 3 different. Each row represents results at different time-steps. The columns are VideoPose, PoseCNN and Ground truth visualisations respectively.}
    \label{fig:qualitative1}
\end{table*}

\begin{table*}[!ht]
\centering
\resizebox{1\textwidth}{!}{%
\scalebox{1}{%
  \begin{tabular}{|c|c|}
    \hline
     & VideoPose \hspace{2in} PoseCNN \hspace{2in}  GT \\ \hline
    t = 1 & \includegraphics[width=1.1\textwidth]{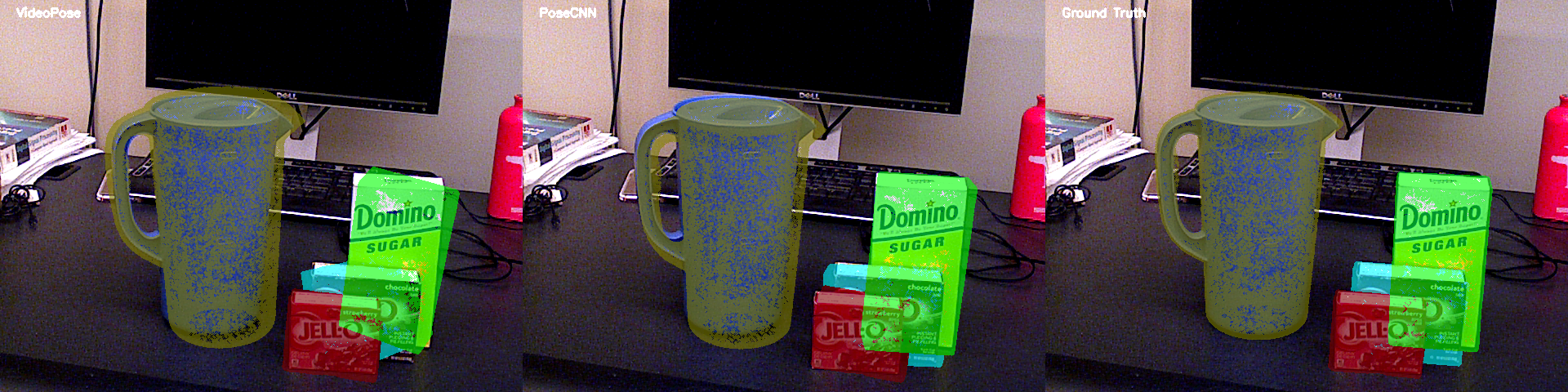} \\ \hline
    t = 2 & \includegraphics[width=1.1\textwidth]{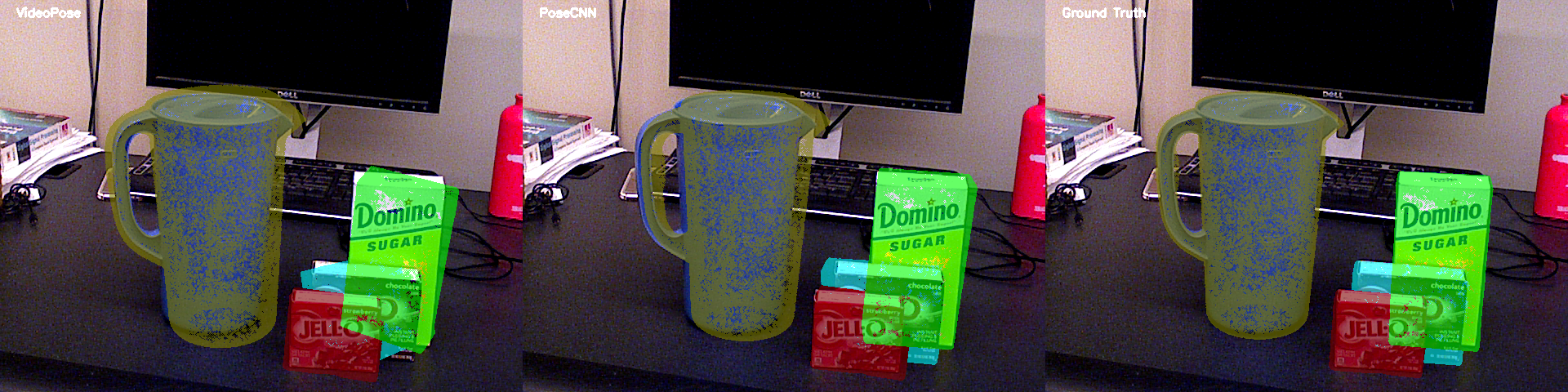} \\  \hline
    t = 3 & \includegraphics[width=1.1\textwidth]{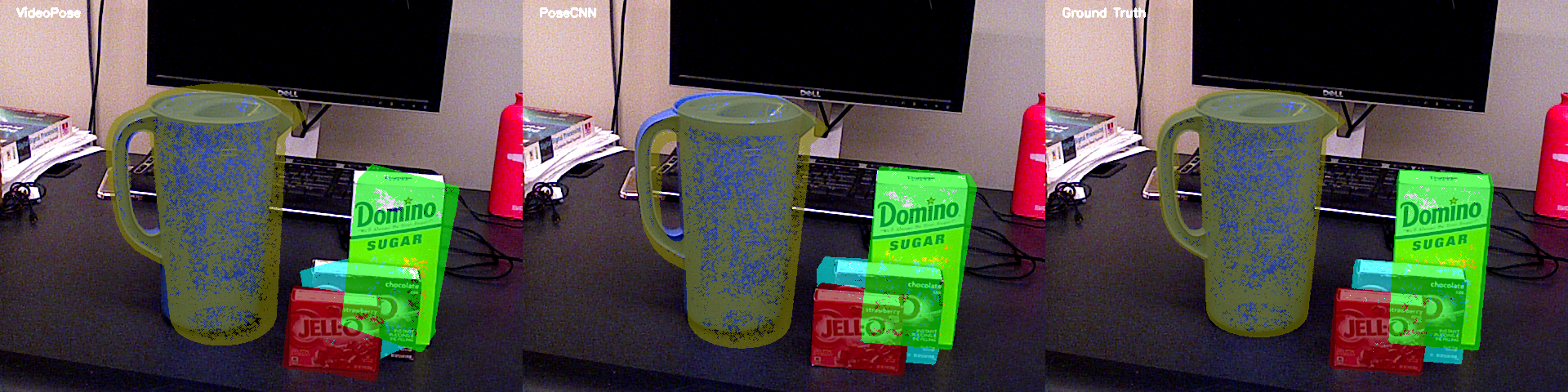} \\ \hline
    t = 4 & \includegraphics[width=1.1\textwidth]{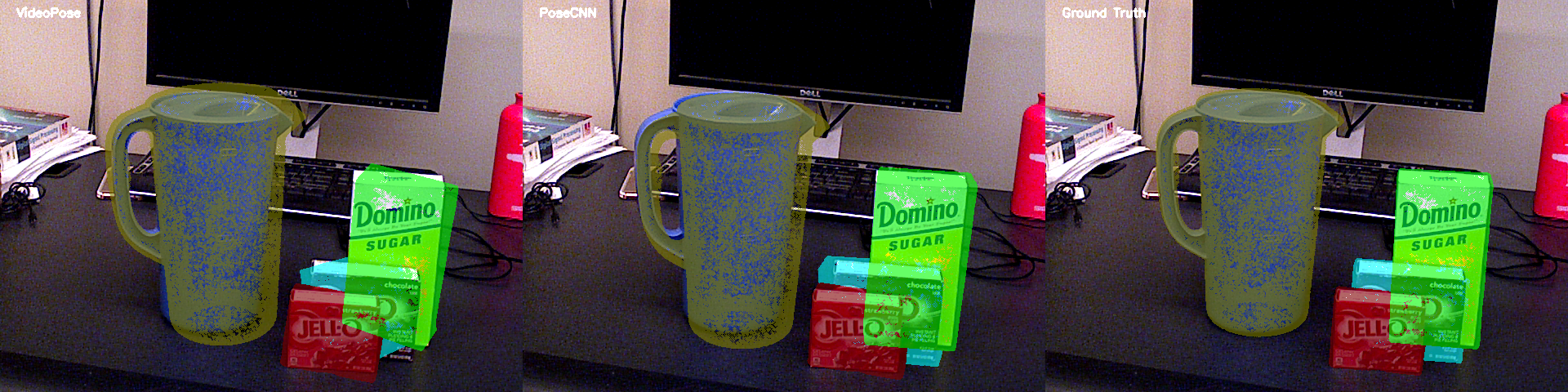} \\ \hline
    t = 5 & \includegraphics[width=1.1\textwidth]{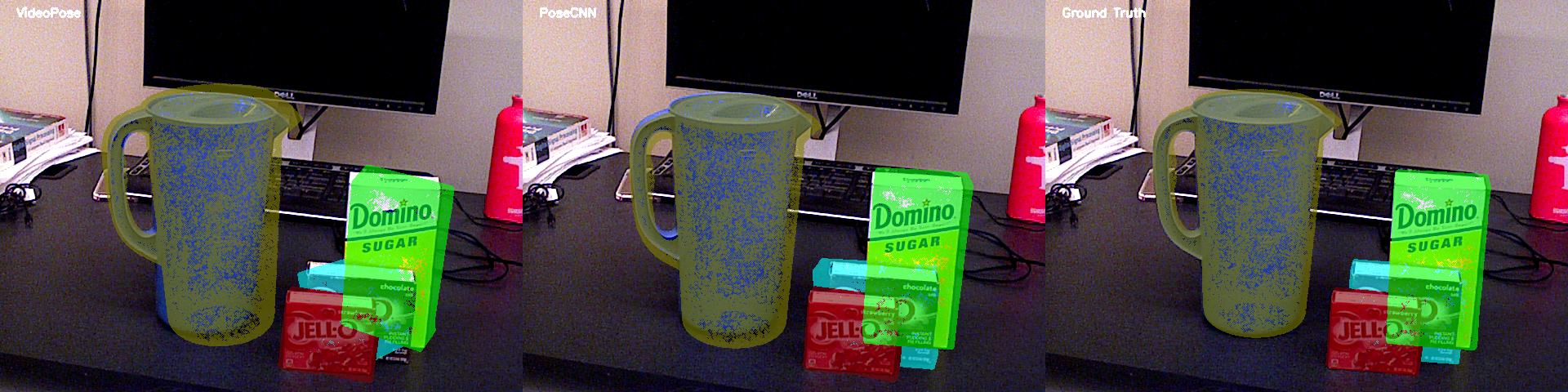}\\ \hline
  \end{tabular}
  }
  }\vspace{.1in}
  \caption{Visualisations of the estimated poses on YCB-Dataset. Each row represents results at different time-steps. The columns are VideoPose, PoseCNN and Ground truth visualisations respectively.}
    \label{fig:qualitative2}
\end{table*}
\section{Conclusion}
\label{sec:conclusion}
In this work, we introduce VideoPose, a simple convolutional neural network architecture to estimate object 6D poses from videos. We demonstrate that by using the $6$D predictions from the previous frames, we can significantly improve $6$D predictions in the subsequent frames. We also conduct an extensive ablation study on different design choices of the network, and show that our model is able to learn and utilise the features from previous predictions regardless of the network choices. Finally, the proposed network performs in real-time at 30fps, thereby improving the time efficiency over previous approaches. As a future work, we would like to further improve our architecture with a better temporal module and model the relationship with the camera transformation and the objects. Our method successfully maintains consistency in pose estimation between frames, however, still depends on the initial frame estimation. We would like to investigate further on improving this, while maintaining the computational efficiency. \\

{
\small
\bibliographystyle{ieee_fullname}
\bibliography{egbib}
}

\clearpage
\appendix
\begin{table*}[!ht]
\centering
\resizebox{0.9\textwidth}{!}{%
\scalebox{1}{%
  \begin{tabular}{|c|c|}
    \hline
     & VideoPose \hspace{2in} PoseCNN \hspace{2in}  GT \\ \hline
    t = 1 & \includegraphics[width=1.1\textwidth]{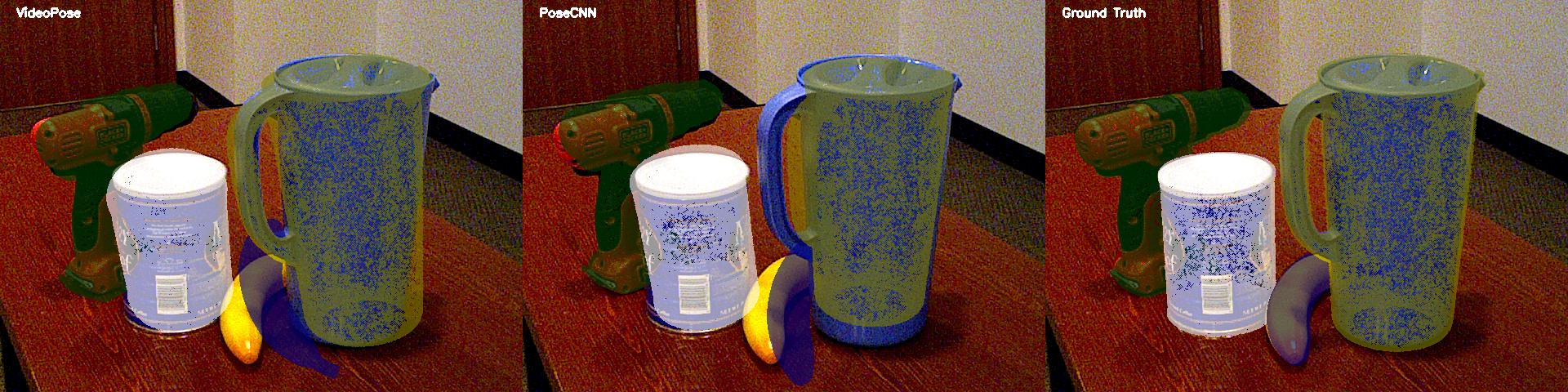} \\ \hline
    t = 2 & \includegraphics[width=1.1\textwidth]{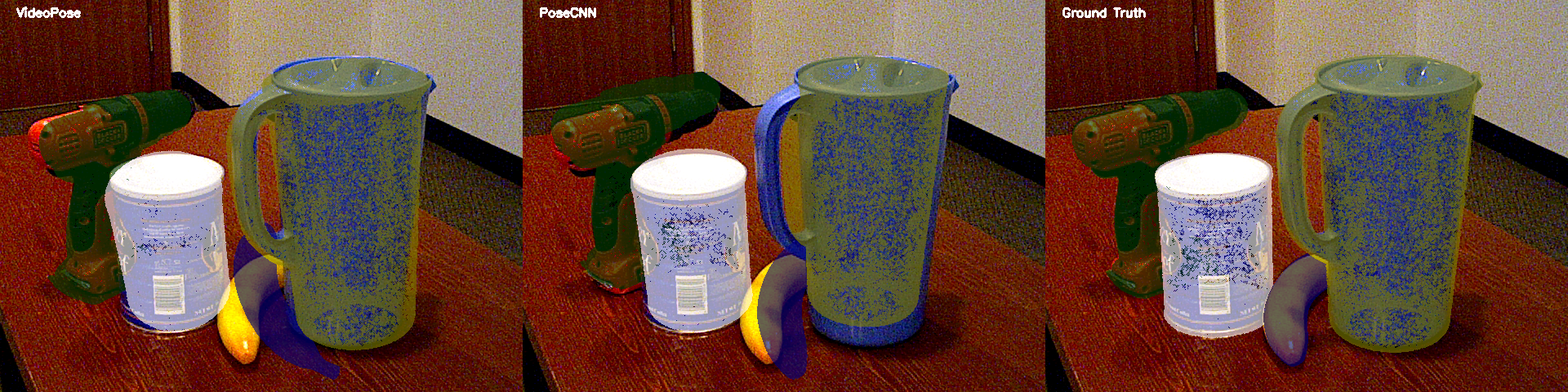} \\  \hline
    t = 3 & \includegraphics[width=1.1\textwidth]{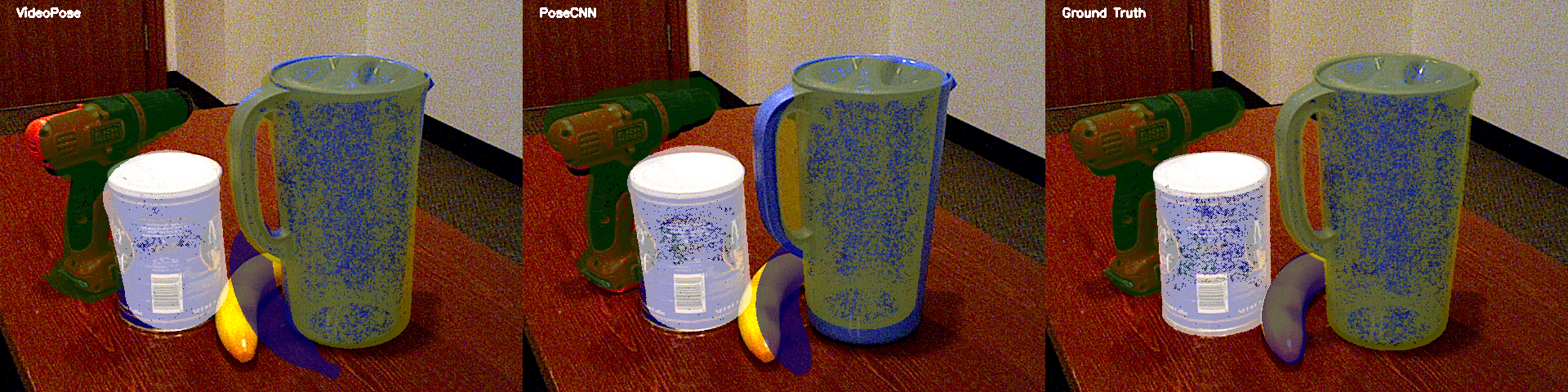} \\ \hline
    t = 4 & \includegraphics[width=1.1\textwidth]{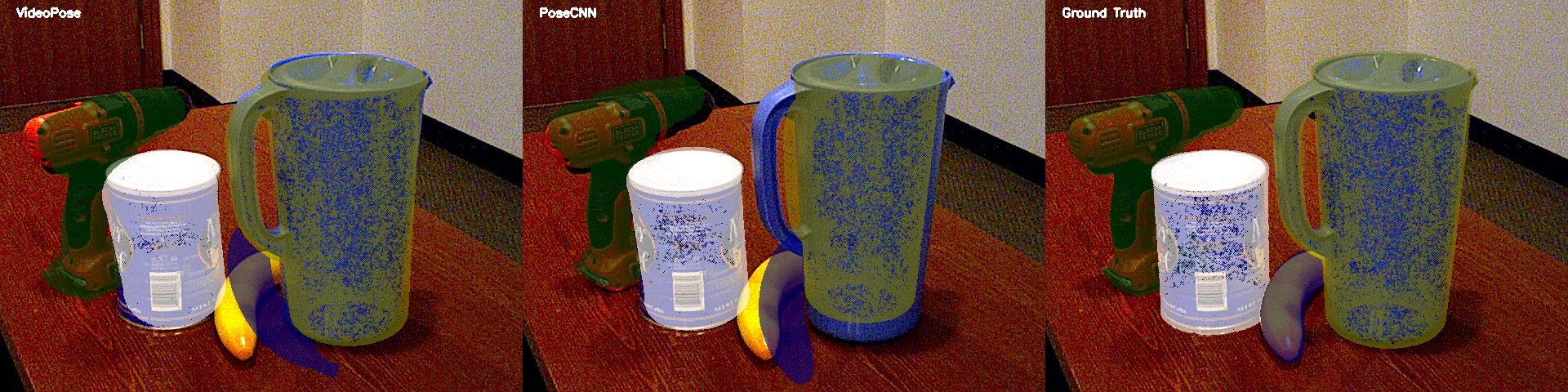} \\ \hline
    t = 5 & \includegraphics[width=1.1\textwidth]{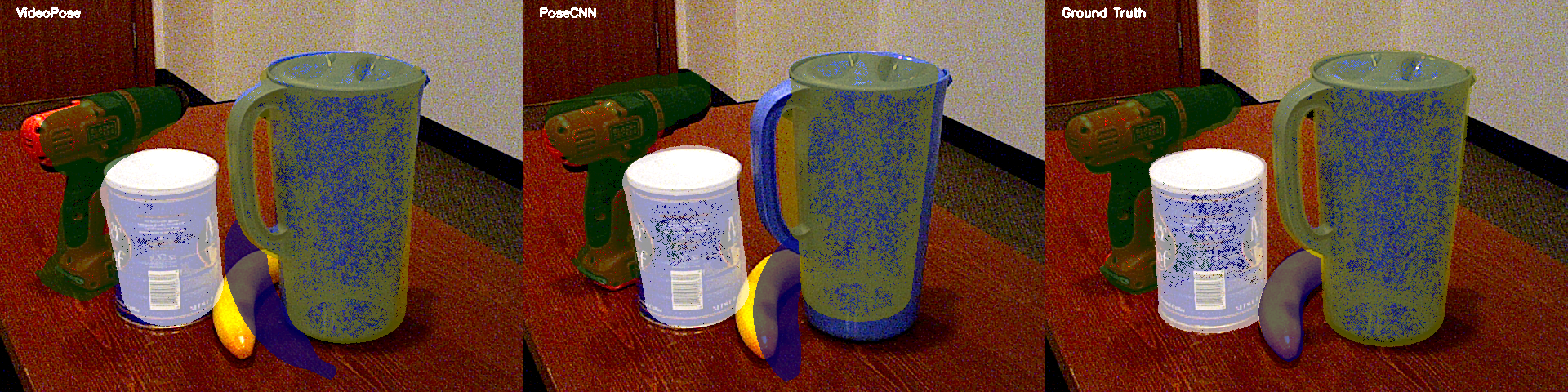}\\ \hline
  \end{tabular}
  }
  }\vspace{.1in}
  \caption{Visualisations of the estimated poses on YCB-Dataset. Each row represents results at different time-steps. The columns are VideoPose, PoseCNN and Ground truth visualisations respectively.}
\end{table*}

\begin{table*}[!ht]
\centering
\resizebox{1\textwidth}{!}{%
\scalebox{1}{%
  \begin{tabular}{|c|c|}
    \hline
     & VideoPose \hspace{2in} PoseCNN \hspace{2in}  GT \\ \hline
    t = 1 & \includegraphics[width=1.1\textwidth]{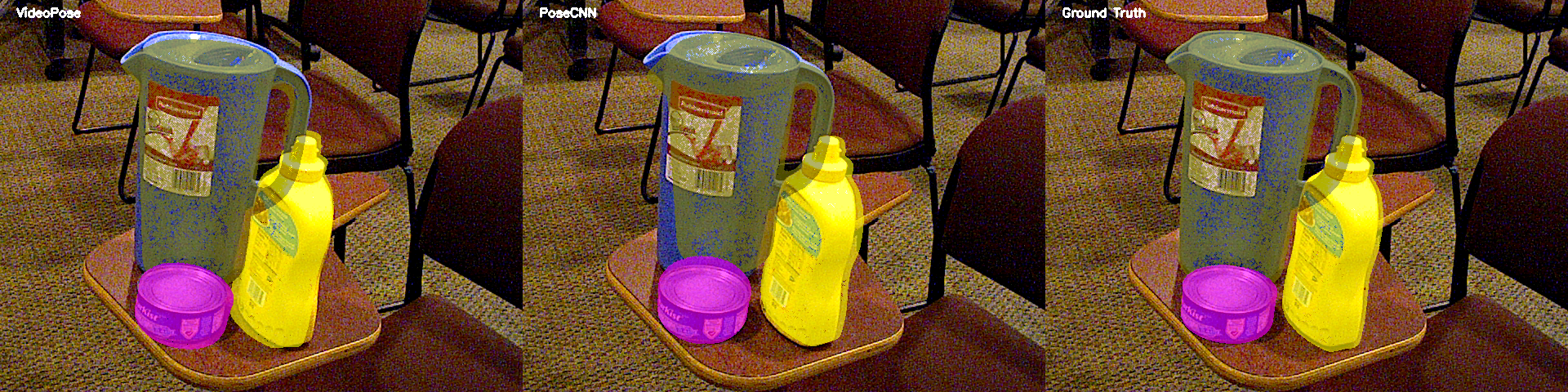} \\ \hline
    t = 2 & \includegraphics[width=1.1\textwidth]{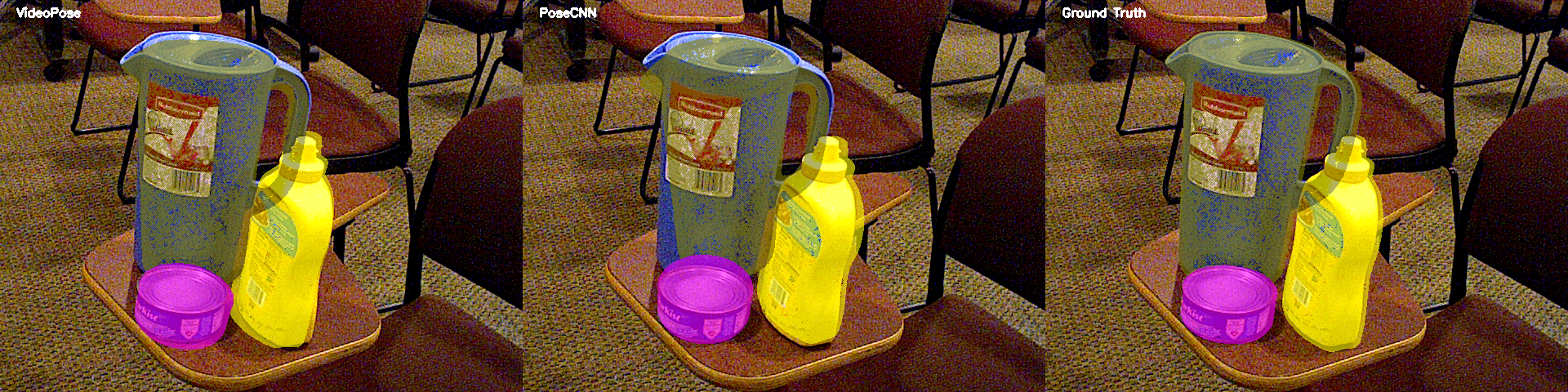} \\  \hline
    t = 3 & \includegraphics[width=1.1\textwidth]{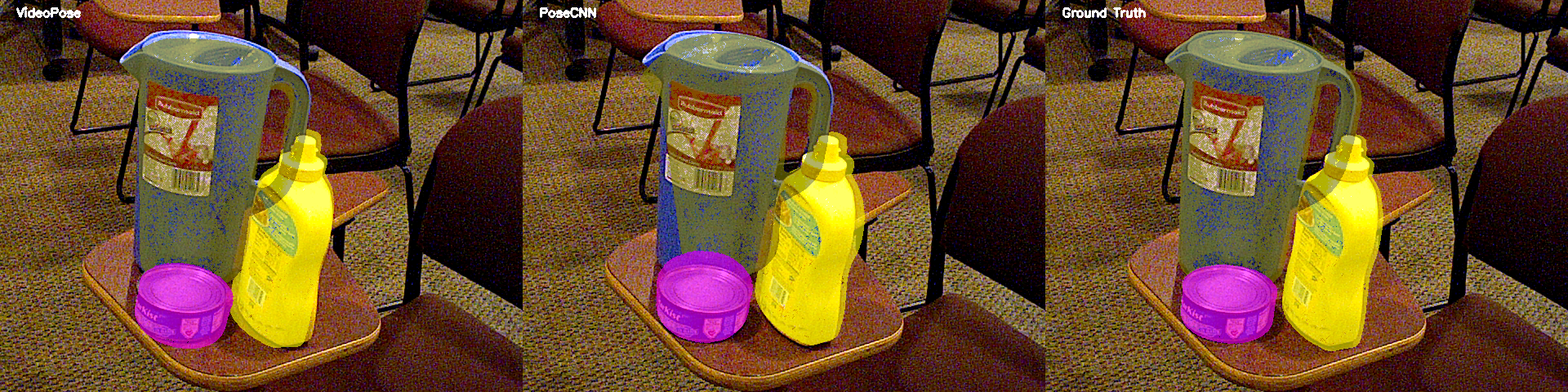} \\ \hline
    t = 4 & \includegraphics[width=1.1\textwidth]{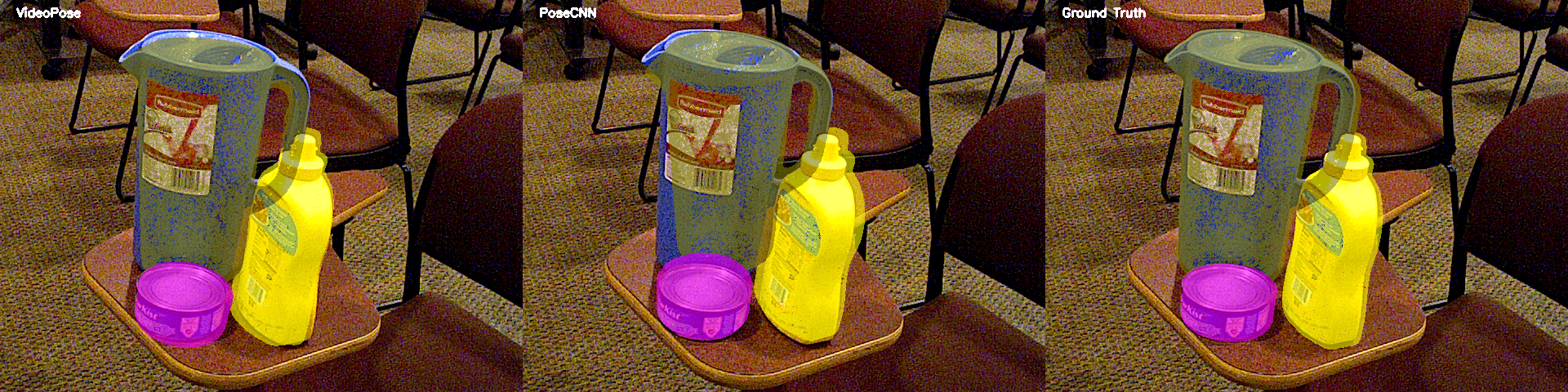} \\ \hline
    t = 5 & \includegraphics[width=1.1\textwidth]{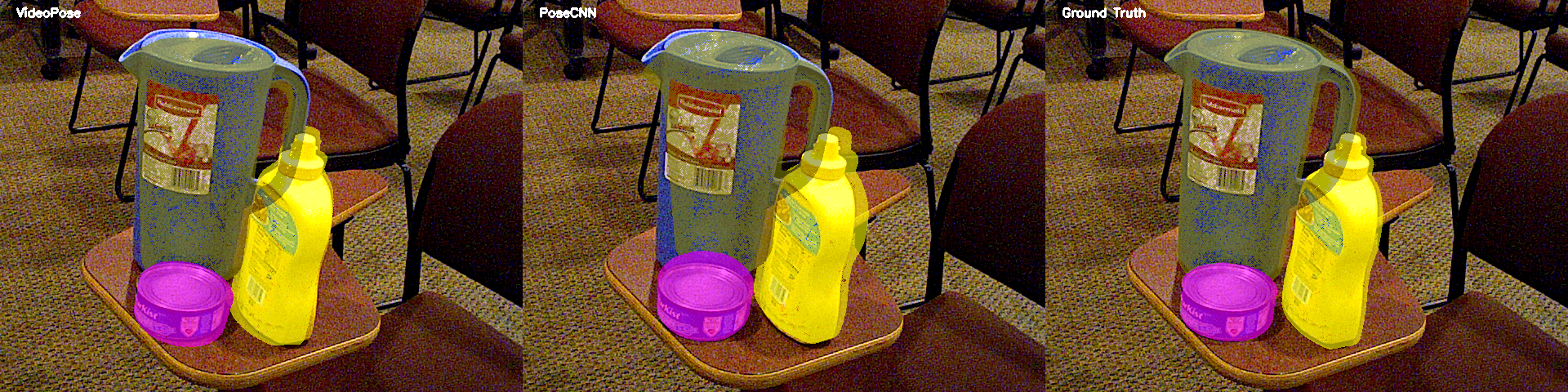}\\ \hline
  \end{tabular}
  }
  }\vspace{.1in}
  \caption{Visualisations of the estimated poses on YCB-Dataset. Each row represents results at different time-steps. The columns are VideoPose, PoseCNN and Ground truth visualisations respectively.}
\end{table*}

\begin{table*}[!ht]
\centering
\resizebox{1\textwidth}{!}{%
\scalebox{1}{%
  \begin{tabular}{|c|c|}
    \hline
     & VideoPose \hspace{2in} PoseCNN \hspace{2in}  GT \\ \hline
    t = 1 & \includegraphics[width=1.1\textwidth]{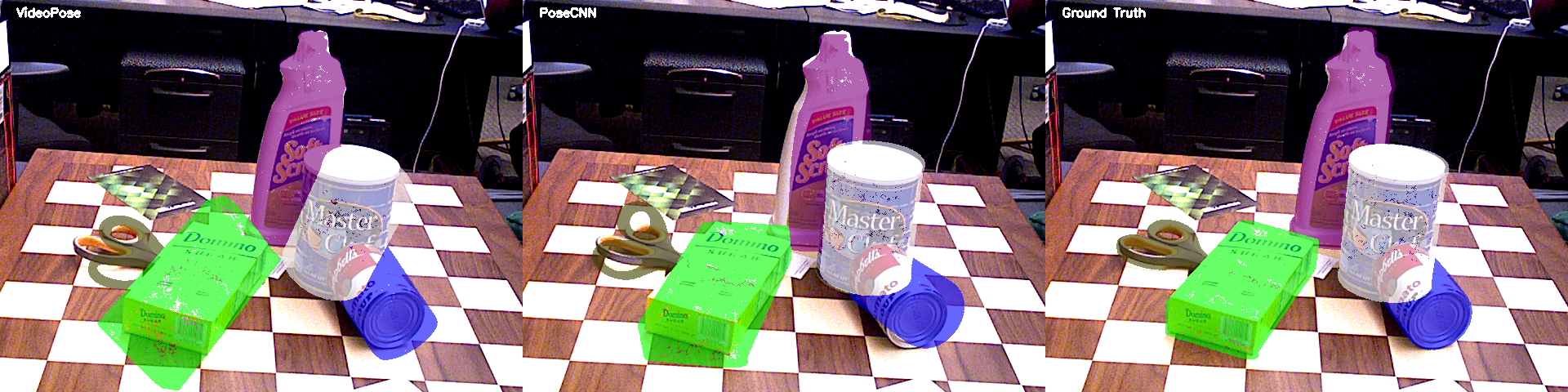} \\ \hline
    t = 2 & \includegraphics[width=1.1\textwidth]{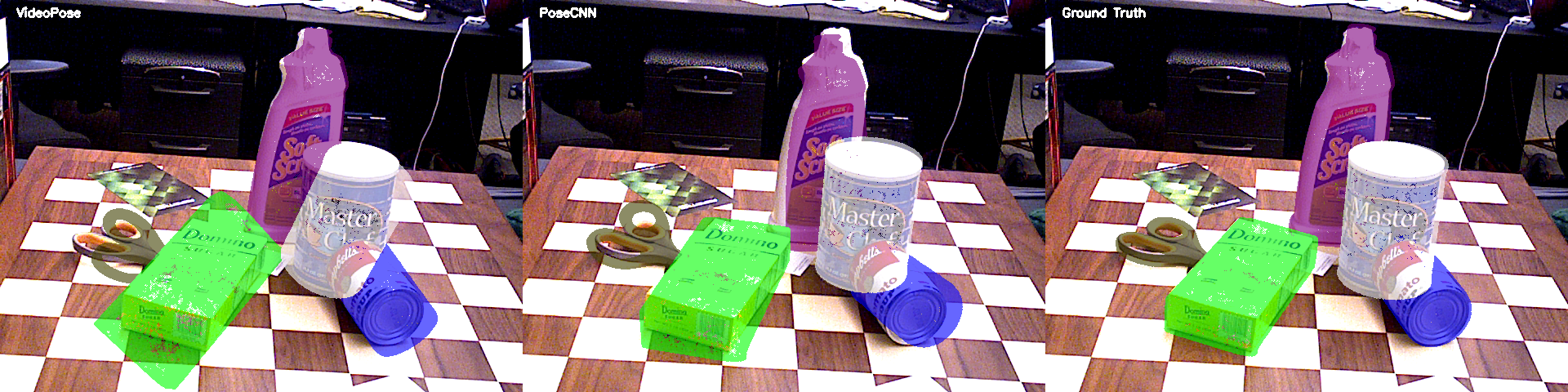} \\  \hline
    t = 3 & \includegraphics[width=1.1\textwidth]{pics/output/keyframe_video_14_0051_000817.png} \\ \hline
    t = 4 & \includegraphics[width=1.1\textwidth]{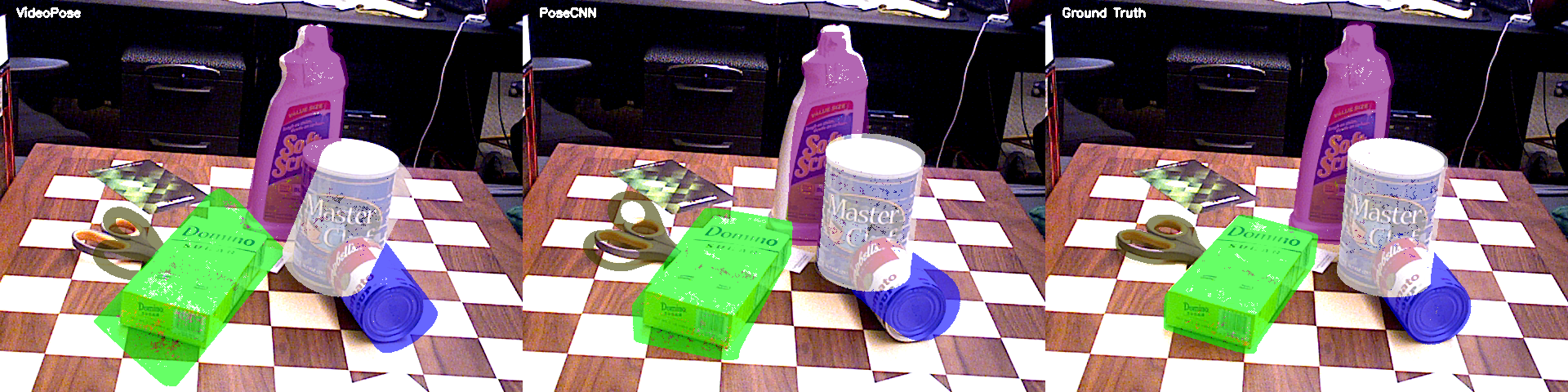} \\ \hline
    t = 5 & \includegraphics[width=1.1\textwidth]{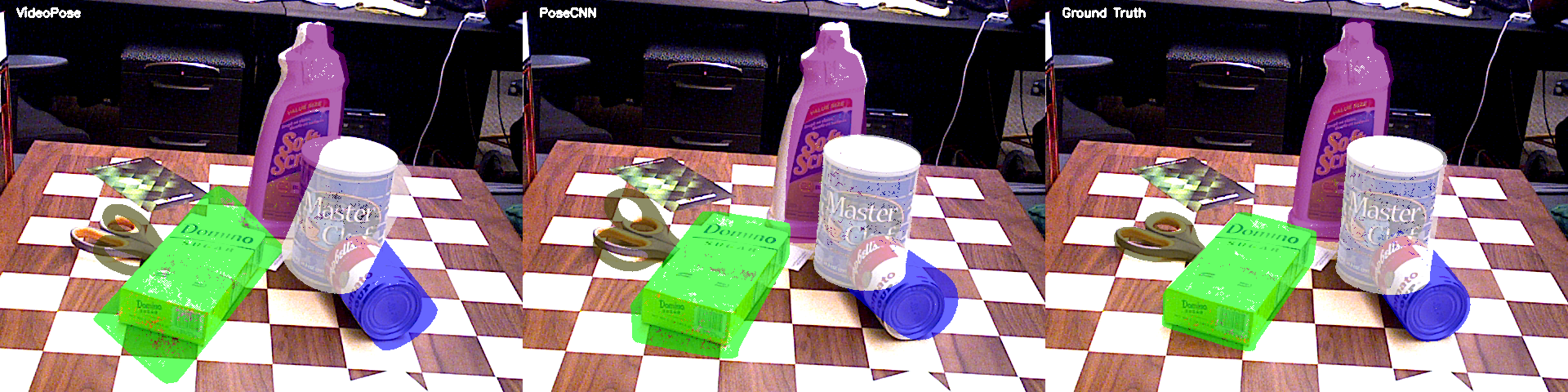}\\ \hline
  \end{tabular}
  }
  }\vspace{.1in}
  \caption{Visualisations of the estimated poses on YCB-Dataset. Each row represents results at different time-steps. The columns are VideoPose, PoseCNN and Ground truth visualisations respectively.}
\end{table*}
\end{document}